\documentclass[runningheads,a4paper]{llncs}
\usepackage{amssymb}
\usepackage{graphicx}
\usepackage[utf8]{inputenc}
\usepackage{url}
\usepackage{courier}
\usepackage{listings}
\newcommand{\keywords}[1]{\par\addvspace\baselineskip
\noindent\keywordname\enspace\ignorespaces#1}
\begin{document}

\mainmatter

\title{Human-Aware Sensor Network Ontology: Semantic Support for Empirical Data Collection}

\author{Paulo Pinheiro\inst{1}\and Deborah L. McGuinness\inst{1}\and Henrique Santos\inst{1,2}}
\authorrunning{Paulo Pinheiro et al.}
\titlerunning{HASNetO: Semantic Support for Empirical Data Collection}

\institute{Rensselaer Polytechnic Institute, Troy, NY, U.S.A. \and Universidade de Fortaleza, Fortaleza, CE, Brazil}

\toctitle{HASNetO: Semantic Support for Empirical Data Collection}
\tocauthor{P. Pinheiro, D. L. McGuinness and H. Santos}
\maketitle

\begin{abstract}
Significant efforts have been made to understand and document knowledge related to scientific measurements. Many of those efforts resulted in one or more high-quality ontologies that describe some aspects of scientific measurements, but not in a comprehensive and coherently integrated manner.  For instance, we note that many of these high-quality ontologies are not properly aligned, and more challenging, that they have different and often conflicting concepts and approaches for encoding knowledge about empirical measurements. As a result of this lack of an integrated view, it is often challenging for scientists to determine whether any two scientific measurements were taken in semantically compatible manners, thus making it difficult to decide whether measurements should be analyzed in combination or not. In this paper, we present the Human-Aware Sensor Network Ontology that is a comprehensive alignment and integration of a sensing infrastructure ontology and a provenance ontology. HASNetO has been under development for more than one year, and has been reviewed, shared and used by multiple scientific communities. The ontology has been in use to support the data management of a number of large-scale ecological monitoring activities (observations) and empirical experiments.

\keywords{empirical data integration, data quality, measurement semantics, HASNetO}
\end{abstract}

\section{Motivation}

Scientific communities are experiencing a significant increase in the availability of empirical data due to the falling cost of sensors along with the growing ease of sensor deployment and with sensor data distribution over the internet. The same communities are also experiencing increasing pressure from a variety of stakeholders to see their empirical data consolidated, analyzed, and used to explain a broad range of unanswered scientific problems. However, this consolidation and analysis presents challenges since scientists are not yet fully equipped to understand the quality and semantics of scientific measurements with the data and often limited annotations typically available today. Many voice a strong need for a comprehensive vocabulary capable of encoding and supporting systematic understanding of metadata about empirical data, which would enable sound integration of empirical data.

We present the Human-Aware Sensor Network Ontology (HASNetO) that is a comprehensive alignment and integration of well-established ontologies for encoding scientific sensing infrastructures, scientific observations, and provenance. The integrated ontology is available at http://hadatac.org/ont/hasneto. Supporting ontologies for HASNetO and previous versions of HASNetO can also be found at http://hadatac.org. A comprehensive infrastructure for managing HASNetO-based knowledge bases, which is not discussed in this paper, is available at https://github.com/paulopinheiro1234/hadatac. One of the immediate benefits of HASNetO is its capability of describing comprehensive knowledge graphs about empirical data.

We have used this graph to systematically annotate and amplify the relevance of scientific measurements stored in database systems, in support of three major projects: Jefferson Project~\cite{jefferson}, Center for Architectural Sciences and Ecology's Build Ecology Program for the City of New York\footnote{http://www.case.rpi.edu/page/academics.php}, and for Smart City activities in Fortaleza, Brazil~\cite{santos_contextual_2015}. Throughout these projects, more than eighty scientists in multiple disciplines are exposed to a new generation of graph-enabled tools for retrieving their data, for retrieving the data from other scientists, for retrieving their data in combination with data from other scientists, and to understand the meaning of data retrieved through complex queries, whether the data has been measured by them or by other scientists.

The rest of this paper is organized as follows. In Section 2, we use a diagram to discuss a typical scenario where empirical data is generated and managed. Section 3 presents a categorization of knowledge related to scientific measurements that is often described as measurement metadata. In Section 4, we introduce the Human-Aware Sensor Network Ontology that provides concepts and relationships used to encode the knowledge discussed in Section 3. In Section 5, we compare our work on HASNetO with other initiatives. A more comprehensive discussion about the current impact of our HASNetO work including future work is described in Section 6. Finally, we summarize our work in Section 7.

\section{A Typical Empirical Data Collection Scenario}

Empirical data are often collected with the use of  instruments that are manually operated by scientists, and sensor networks that are automatically operated but that are still deployed, calibrated and maintained manually.

\begin{figure}
\centering
\includegraphics[height=8cm]{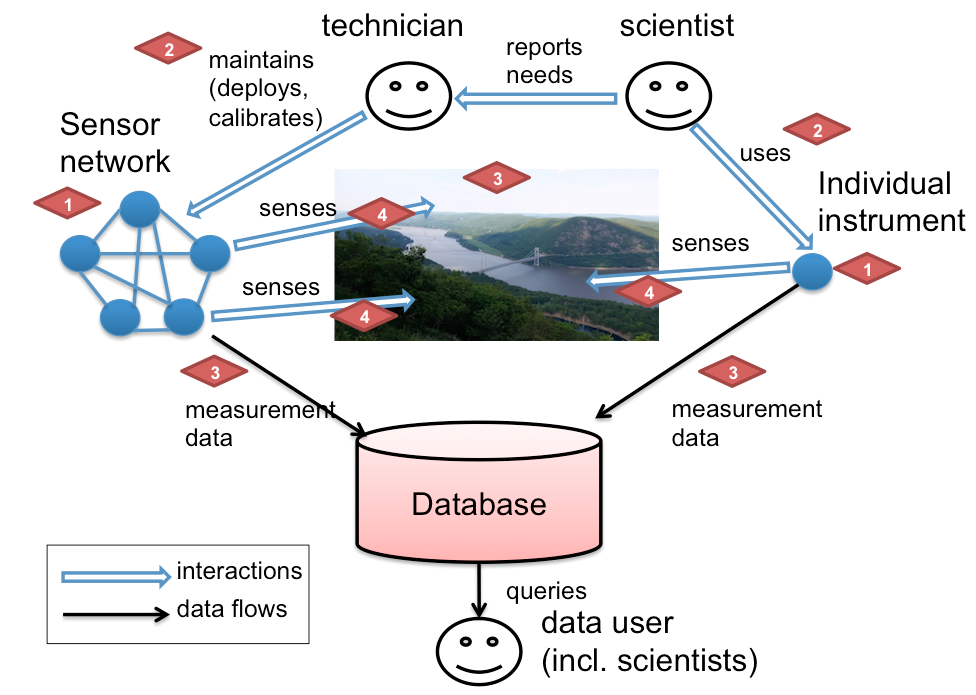}
\caption{Typical database-based data collection infrastructure.}
\label{fig01-new}
\end{figure}

The three faces depicted in Fig.~\ref{fig01-new} represent \textit{human roles} (or just {\em roles}) in a data collection scenario. The {\em Scientist role} is connected to a {\em Technician} role representing the fact that scientists interact with technicians to communicate their needs in terms of how sensor networks are required to be set up and maintained. Human roles in Figure 1 can be performed by any combination of people and roles.

\section{Knowledge Behind Measurement Data}

Figure~\ref{fig01-new} contains ten diamonds that label knowledge related to typical data collection scenarios. The knowledge identified by these diamonds, which is explained in this section, is often captured and recorded as metadata for the empirical data. One of the assumptions behind such recording  is that collected metadata enable data understanding without the need of any explanation from scientists directly involved with the data collection. There are four categories of diamonds; each of which is represented by the number inside the diamonds. Below we describe each of these four diamond categories.

\subsection{Available Measurement Infrastructure}

In Figure~\ref{fig01-new}, Diamond Category ``1'' represents knowledge about available measurement infrastructure. Scientists conceptually understand the configuration and capabilities of measurement infrastructures, including which instruments and detectors are available to them, which platforms these instruments can be or are deployed to, where stationary platforms are located, which paths are taken by mobile platforms, and what physical, chemical, biological and sociological properties the sensors are capable of measuring. Assuming that the knowledge about instruments and sensor networks may affect empirical data  understanding, scientists are expected to share their measurement infrastructure knowledge by encoding such knowledge as measurement data's metadata.

\subsection{Calibrations, Configurations and Deployments of Instruments and Detectors}

In Figure~\ref{fig01-new}, Diamond Category ``2'' represents knowledge about a broad range of human interventions that may affect the quality of measurement data. Knowledge about measurement infrastructure is not nearly enough to explain data generated by instruments in such infrastructures. For instance, many are the factors/events that may affect the way measurements are performed, which are not included in the knowledge about the measurement infrastructure itself. When scientists are operating scientific instruments in isolation, it is evident the importance of documenting how the instruments were operated. More challenging is the process of explaining human interventions in sensor networks, which is often regarded as an automated infrastructure for the collection of scientific data. For example, a badly deployed instrument, e.g., an instrument that is not properly attached to the surface of the deploying platform, can create measurements that are off by a fixed amount, or even worse, that may not be able to execute any measurement, e.g., because the chord providing power to the instrument is not properly connected.

\subsection{Scientific Annotations of Measurements}

In Figure~\ref{fig01-new}, Diamond Category ``3'' represents knowledge about what has been measured, and how the measurement has been represented in terms of units. When measurements occur, these are measurements of physical, chemical, biological, cultural, and social properties of so-called entities of interests. For example, using Air as an entity of interest, we can say that the air temperature is a physical property of air and that the CO$_2$ concentration of the air is a chemical property of the same entity of interest. For data understanding, it is important for one to know the properties are that are being measured, e.g., temperature and CO$_2$ concentration, and what entities of interest are behind these properties, e.g., air. Moreover, it is important to understand the unit used to represent the measurements and the semantic context, (e.g., air is ‘outside air’ as opposed of air inside of a room, a lab or a shelter) that may affect, for instance, the actual measurement of both air temperature and air concentration of CO$_2$.

\subsection{Provenance of Sensor Network Activities}

In Figure~\ref{fig01-new}, Diamond Category ``4'' represents knowledge about the provenance of both human interventions, as well as of each measurement. For each measurement, it is important to know when and where the measurement was done. What was the combination of sensing devices used to support the measurement? Was any configuration parameter provided to the sensing devices to allow the devices to operate the way they were operating at the time the measurements were done?

\section{HASNetO: The Human-Aware Sensor Network Ontology}

HASNetO aims to provide the concepts and vocabularies needed to encode empirical data’s metadata as identified and described in Section 3. HASNetO is built on top of three ontologies that were integrated and extended under the single name of HASNetO: The Extensible Observation Ontology (OBOE)~\cite{oboe}, the Virtual Solar Terrestrial Observatory (VSTO)\footnote{The vstoi namespace refers to the instrument portion of the VSTO ontology family.}~\cite{vsto}, and the World Wide Web's Provenance Ontology (PROV-O)~\cite{prov}.

\subsection{Encoding Knowledge about Sensor Networks and Individual Instruments}

\begin{figure}
\centering
\includegraphics[height=3.5cm]{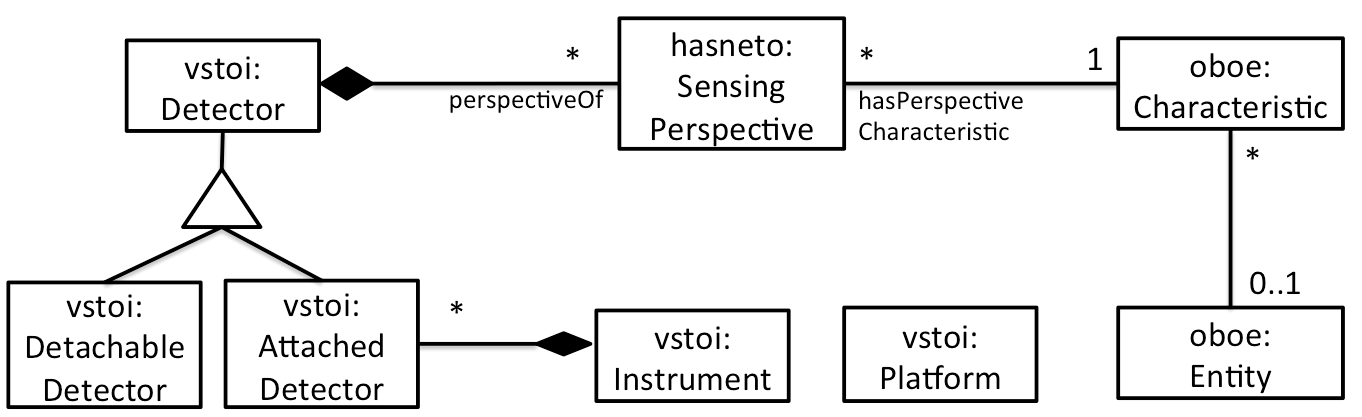}
\caption{HASNetO concepts describing measurement infrastructure.}
\label{fig03}
\end{figure}

HASNetO contains content related to sensor networks although it does not have a Sensor concept. We observe that the term sensor is used to refer to detectors, instruments, and often to combinations of detectors and instruments. To avoid further confusion, HASNetO advocates for the use of the terms detectors and instruments knowing that it may be difficult to perceive which part of a device is a detector (or detectors) and which part is an instrument. For instance, a thermometer may include an embedded, non-detachable detector. HASNetO breaks down the elements of measuring infrastructures into three categories, as shown in Fig.~\ref{fig03}.

\begin{itemize}
\item \textit{vstoi:Platform}: An object that keeps the instrument in a specific location to ensure that it is recording data about the selected location. A platform may also provide overhead services, such as providing power to the instrument and a data connection. Sometimes a platform is mobile like a plane or a person, or stationary like a tower of a weather station.
\item \textit{vstoi:Instrument}: n object that receives sensed signals from detectors and processes these signals into numerical values. For example, consider a tipping bucket rain gauge. Inside the tipping bucket rain gauge is a magnet-based detector that detects when the bucket tips. However, in order for this signal to be meaningful, the detector needs a bucket with a known diameter, a funnel to direct water into the bucket, etc. Together, these make up the instrument.
\item \textit{vstoi:Detector}: An object that it is capable of sensing environmental properties by collecting physical signals about these properties, translating these physical signals into (most often electrical) signals, and forwarding these electrical signals to instruments. Transducer is another name for detector. Detector metadata are collected because detectors may be interchangeable, that is they can be removed from one instrument and plugged into another.
\item \textit{vstoi:Deployment}: An activity of physically deploying an instrument and its attached detectors to a platform. This activity indicates that a single instrument is ready to start collecting data.
\end{itemize}

For more sophisticated devices, detectors and instruments are sometimes available as distinct hardware components, and thus easier to be mapped into HASNetO concepts. For ordinary instruments, it may be appropriate to make explicit the existence of attached detectors since properties like measurement accuracy and measurement ranges, which are detector’s properties, are not listed as instrument properties.

Fig.~\ref{fig03} also shows that OBOE provides concepts for describing entities of interest and their measured properties. More specifically, measurements are of properties of entities of interests. These measured properties are called oboe:Characteristics. These oboe terms are listed below along with their original definitions.

\begin{itemize}
\item oboe:Entity “denotes a concrete or conceptual object that has been observed (e.g., a tree, a community, an ecological process).”
\item oboe:Characteristic “represents a property of an entity that can be measured (e.g., height, length, or color).”
\end{itemize}

\subsection{Encoding Knowledge about Measurements}

Imagine two data sets of “air temperature" measurements obtained from a common weather station thermometer and using Celsius to represent measured values. These measurements still could use different hardware and software configurations or calibrations for the platform and observing agent – in this case the weather station and the thermometer respectively, thereby making the measurements difficult to compare or use in combination. For example, during one use of the thermometer, it was calibrated to operate in the [0$^o$,20$^o$] range when the actual temperature was in the operation range. During another use of the thermometer, it was still calibrated to operate in the [0$^o$,20$^o$] range although the actual temperature was in the [-10$^o$,10$^o$]. As a result of a bad calibration decision, the thermometer ended up generating data that may be classified as of low quality. OBOE is aware of the impact of context in observation data management, which is why the ontology provides a context concept. The notion of context in OBOE provides a start for encoding context, however it does not include descriptions of what constitutes a context property, and more importantly, what does not constitute a context property.

\begin{itemize}
\item oboe:Measurement is an assertion that a characteristic of an   entity was measured and/or recorded. A measurement is also composed of a value, a measurement standard, and a precision (associated with the measured value). Measurements also encapsulate characteristics that were recorded, but that were not necessarily measured in a physical sense. For example, the name of a location and a taxon can be captured through measurements.
\item oboe:Standard defines a reference for comparing or naming entities via a measurement. A standard can be defined intentionally (e.g., as in the case of units) or extensionally (by listing the values of the standard, e.g., for color this might be red, blue, yellow, etc).
\item hasneto:DataCollection defines the technical activity of the collection of data that is empirically observed. So far, the state of the art of semantics for observations and measurements characterizes this activity as an Observation. The HASNetO ontologies take the position that the concept of Observation is a scientific activity while most if not all existing ontologies embody the position that describe the technical activity of data collection.
\item oboe:Observation represents an ‘observed entity’ that is, an entity that was observed by an observer. An observation often consists of measurements that refer to one or more measured characteristics of the observed entity.
\end{itemize}

\begin{figure}
\centering
\includegraphics[height=5cm]{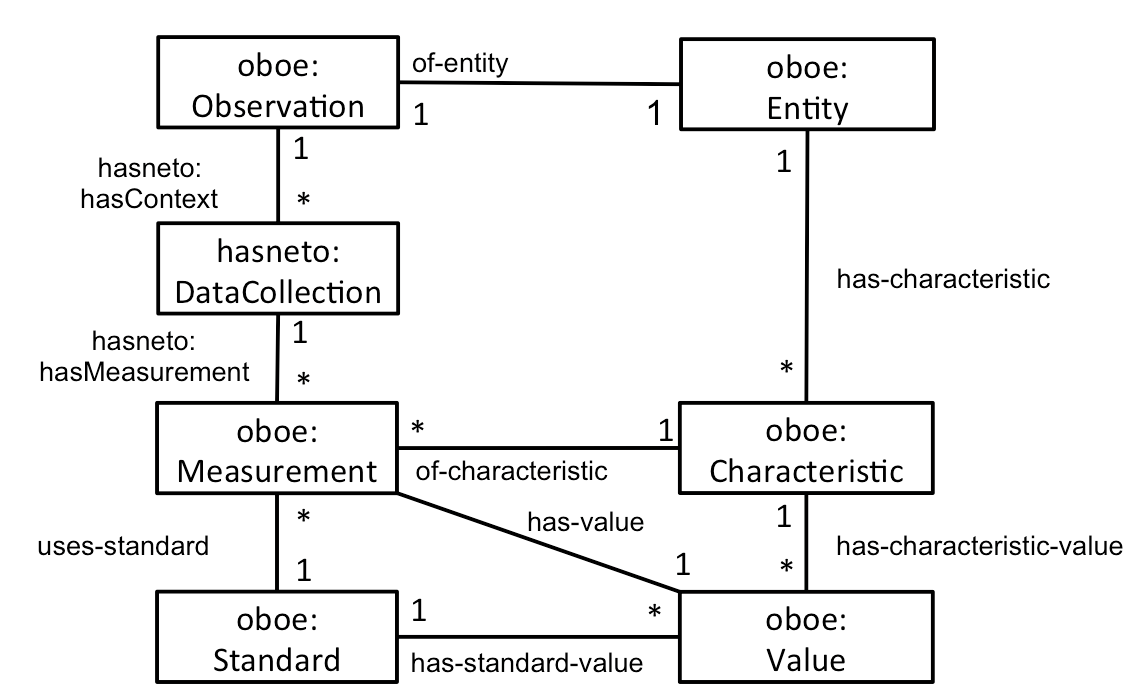}
\caption{HASNetO concepts describing scientific measurements.}
\label{fig04}
\end{figure}

\subsection{Encoding Knowledge about Human Interventions}

Provenance knowledge is an important part of contextual knowledge that is often not fully captured in many scientific applications. HASNetO is a major beneficiary of all the previous work developed by the provenance community in defining a truly general-purpose vocabulary for provenance, which is the W3C PROV language~\cite{prov,provxg}. In terms of empirical data, we use provenance any time we have technical activities in support of scientific activities that may affect measurement data. For example, humans are often heavily involved in technical activities such as instrument deployments, platform maintenance, instrument and detector's calibration and soon. This human involvement in the scientific process often is not encoded, yet it can impact measurements and their interpretations.

\begin{figure}
\centering \includegraphics[height=5cm]{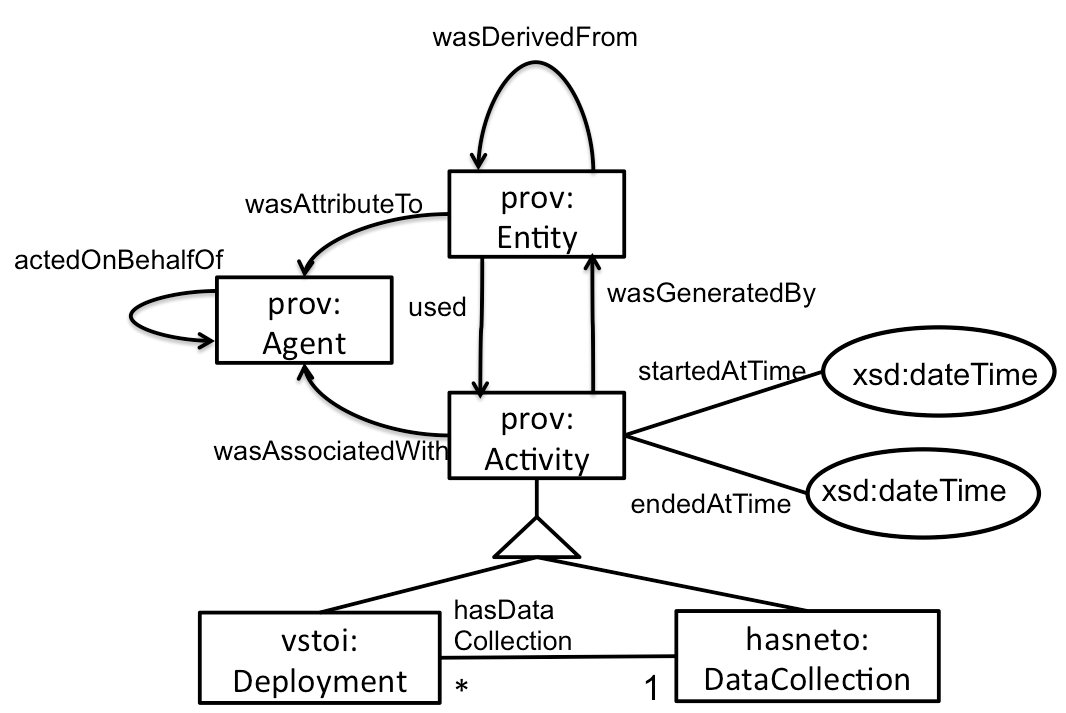}
\caption{HASNetO concepts describing the provenance of technical
activities in support of empirical data collections.}
\label{fig05}
\end{figure}

Fig.~\ref{fig05} shows how vstoi:Deployment and hasneto:DataCollection, which are two of the most important technical activities related to empirical data, are defined as subclasses of prov:Activity. These two subclasses of prov:Activity have been discussed previously. Below, we briefly describe prov:Activity and its two complementary classes prov:Agent and prov:Entity.

\begin{itemize}
\item prov:Activity is “how PROV entities come into existence and how their attributes change to become new entities, often making use of previously existing entities to achieve this.”
\item prov:Agent “takes a role in an activity such that the agent can be assigned some degree of responsibility for the activity taking place. An agent can be a person, a piece of software, an inanimate object, an organization, or other entities that may be ascribed responsibility. When an agent has some responsibility for an activity, PROV says the agent was associated with the activity, where several agents may be associated with an activity and vice-versa.” In HASNetO terms, we see that some prov:Activity instances in support of data collection are mainly performed by humans while others are mainly performed by machines. However, it can be challenging and in fact unnecessary to classify these activities as long we can fully describe the exact involvement of each agent in each activity, including the fact that the agent is a human or a machine.
\item prov:Entity is defined as “physical, digital, conceptual, or other kinds of thing.” In HASNetO, prov:Entity is used to represent, for instance, samples that have been collected and that are going to be further analyzed in a lab, that is where scientific measurements and data collections occur. Also, prov:Entity is used to specify any information that is fed into an platform, instrument or detector that change the behavior of any of these measuring devices. Finally, while an instance of prov:Entity may be an instance of oboe:Entity and vice-versa, we prefer to treat them separate considering their distinct roles in scientific activities.
\end{itemize}

\section{Related Work}

Ongoing research activities in support of semantic sensor networks make use of the description of instruments and detectors (many times called just ``sensors'' in the literature) to maintain complex networks of sensors, while providing integration of the collected data. In \cite{compton_survey_2009}, twelve different sensor network ontologies are studied and compared.  The authors  concluded that no ontology (or combination of ontologies) at that time was able to describe properties required for the stipulated capabilities of sensor networks. This work preceded the W3C's Semantic Sensor Network Ontology (SSN) \cite{compton_ssn_2012}. SSN is an ontology that aims to describe sensors, observations and related concepts, like sensor capabilities, measurement processes and deployments. SSN provides vocabulary capable of annotating data in a manner that makes it possible to determine if data are coming from a certain sensor, and  if they are using some specific process to measure a certain property of an entity of interest. BOnSAI \cite{stavropoulos_bonsai:_2012} and SESAME Meter Data Ontology \cite{fensel_sesame-s:_2012} are other sensor network ontologies that are focused on smart buildings. Despite the capability of describing the tracking of single measurements, those ontologies are not concerned with the linking of measurements to units or entities of interest. Although the ontologies mentioned above are capable of describing sensor networks used to collect data, SSN does not rely on standard provenance approaches, like the W3C's PROV, and thus are limited when they attempt to describe human interventions to sensor networks. Besides that, the SSN ontology does not provide any software framework describing how the vocabulary should be used to enable management of empirical data. BOnSAI and SESAME are not scientific centric ontologies. They are unable to track human interventions to the network by means of deployments, calibrations or sensor settings, and are also unable to explain the implication of these interventions on empirical data quality.

The concept of Observation data is treated in the literature \cite{quine_stimulus_1995} \cite{stasch_stimulus-centric_2009} \cite{probst_ontological_2006} as data that are obtained while sensing some property of an entity from the real world. The result of an observation is a value for that property \cite{usbeck_combining_2014}. Content annotation is crucial when dealing with observation data (do they talk about data quality, and more specifically, how to differentiate measurements when they are from distinct data collections, i.e., distinct calibrations, setting, etc.?). It enables some level of interoperability and discoverability, making the data easier to be used. To leverage this potential, several approaches exist to both model the infrastructure that generates the data and to describe data content and context.

O\&M \cite{cox_observations_2011} is an XML implementation from the Open Geospatial Consortium (OGC) that defines a schema for modeling observations and their results. In \cite{kuhn_functional_2009}, an observation and measurement ontology is proposed that makes use of OGC's definitions. OBOE (The SEEK Extensible Observation Ontology) is an ontology focused primarily on ecology that provides a data model that can capture measurement semantics and that can be used to streamline data integration. To achieve this goal, the OBOE ontology contains concepts and relationships for describing observational datasets.

In other initiatives to annotate scientific data, VSTO provides a data framework for ontology based discovery of datasets across the fields of solar physics, space physics and solar-terrestrial physics from multiple repositories.

\section{Discussion}

\subsection{Systematic Evaluation of HASNetO by Scientists}

One strength of HASNetO comes from the fact that OBOE, PROV and VSTO are mature community-developed ontologies. For instance, OBOE was initiated by an NSF-funded project and has evolved through a number of sponsored research projects. PROV is a recommended standard from W3C endorsed by academic organizations and industry. VSTOI is a by-product of the VSTO ontology~\cite{vsto} which was funded by NSF and NASA awards and, has been influential in the development of Woods Hole’s BCO-DMO Ontology currently used by a large oceanographic community~\cite{bcodmo}.

The HASNetO ontology may be regarded as suited for the management of empirical scientific data if the final querying and browsing capabilities of a HASNetO-based infrastructure is regarded as useful by a community of scientists addressing some data management challenges described in terms of use cases. Our HASNetO ontology powers our prototype HADataC (Human-Aware Data Collection) Framework, which is under development, and has been deployed in support of three major research projects/organizations:

\begin{itemize}
\item At the RPI Tetherless World Constellation in support of the   Jefferson Project developed in collaboration between IBM, Rensselaer Polytechnic Institute (RPI), and The FUND for Lake George~\cite{jefferson};
\item At RPI's Center for Architect, Science and Ecology in support of large empirical observations and experiments in the areas of urban ecology;
\item At the Universidade of Fortaleza’s Smart City Center where scientific observations are conducted to understand the use of city’s resources in support of mass transportation.
\end{itemize}

\subsection{Future Work}

The Human-Aware Science Ontologies (HAScO) is a family of ontologies. HAScO itself is a high-level ontology that describes scientific activities along with supporting technical activities. Within HAScO, data collections are defined as technical activities in support of empirical and simulated data. HASNetO is the HAScO ontology that provides a vocabulary for encoding knowledge about empirical data collection. One overarching goal (and challenge) for HAScO is to provide a vocabulary small enough that domain scientists are comfortable using it, but still rich enough for use in explaining complex relationships involved in the combined used of empirical and computational scientific activities.

\section{Conclusions}

The Human-Aware Sensor Network Ontology (HASNetO) was established as an integrated and comprehensive vocabulary for encoding knowledge related to scientific measurements and their derived empirical data. HASNetO aligns and resolves conflicts from the integration of three community-developed and community-maintained ontologies for observation, sensing, and provenance. Contributions include the identification of appropriate covering ontologies, the alignment between them, the gap analysis, and the gap filling. One key gap that HASNetO  filled relates to providing terms for modeling human interventions related to empirical activities. Sensor deployment and data collection are examples of such human interventions. The exact interpretation of the Observation concept from the OBOE Ontology, and its meaning in terms of data collection was clarified with the creation of a HASNetO concept called {\em DataCollection}.  This is the actual act of collecting data in the context of scientific activities such as an OBOE Observation itself and empirical experiments. It is also worth mentioning that HASNetO clarifies the use of the term ``sensor'' in its description of a sensing infrastructure, and when it is compared against competing
efforts.

Finally, a full explanation of human interventions in measurements generating empirical data is provided by the provenance of empirical data, which is defined as a result of combinations of activities such as VSTO Deployment and HASNetO Data Collection. Moreover, VSTO Deployment and HASNetO Data Collection are defined as PROV Activity's specializations.

\subsubsection{Acknowledgements.} The third author is supported by  CNPq - Brazil - Science Without Borders scholarship.

\bibliography{biblio}

\begin{thebibliography}{10}
\providecommand{\url}[1]{\texttt{#1}}
\providecommand{\urlprefix}{URL }

\bibitem{bcodmo}
Chandler, C., Fox, P., Maffei, A., Alison, M., Groman, R., West, P., Zednik,
  S.: Evolving the bco-dmo search interface-experience with semantic and smart
  search. In: EGU General Assembly Conference Abstracts. vol.~12, p. 14621
  (2010)

\bibitem{compton_ssn_2012}
Compton, M., Barnaghi, P., Bermudez, L., García-Castro, R., Corcho, O., Cox,
  S., Graybeal, J., Hauswirth, M., Henson, C., Herzog, A., Huang, V., Janowicz,
  K., Kelsey, W.D., Le~Phuoc, D., Lefort, L., Leggieri, M., Neuhaus, H.,
  Nikolov, A., Page, K., Passant, A., Sheth, A., Taylor, K.: The {SSN} ontology
  of the {W}3c semantic sensor network incubator group. Web Semantics: Science,
  Services and Agents on the World Wide Web  17,  25--32 (Dec 2012)

\bibitem{compton_survey_2009}
Compton, M., Henson, C., Lefort, L., Neuhaus, H., Sheth, A.: A {Survey} of the
  {Semantic} {Specification} of {Sensors}. CEUR Workshop Proceedings pp. 17--32
  (Oct 2009)

\bibitem{cox_observations_2011}
Cox, S.: Observations and {Measurements} - {XML} {Implementation} (Mar 2011)

\bibitem{fensel_sesame-s:_2012}
Fensel, A., Tomic, S., Kumar, V., Stefanovic, M., Aleshin, S.V., Novikov, D.O.:
  {SESAME}-{S}: {Semantic} {Smart} {Home} {System} for {Energy} {Efficiency}.
  Informatik-Spektrum  36(1),  46--57 (Dec 2012)

\bibitem{vsto}
Fox, P., McGuinness, D.L., Cinquini, L., West, P., Garcia, J., Benedict, J.L.,
  Middleton, D.: Ontology-supported scientific data frameworks: The virtual
  solar-terrestrial observatory experience. Computers \& Geosciences  35(4),
  724--738 (2009)

\bibitem{provxg}
Gil, Y., Cheney, J., Groth, P., Hartig, O., Miles, S., Moreau, L., da~Silva,
  P.P., et~al.: Provenance xg final report. Final Incubator Group Report
  (2010)

\bibitem{kuhn_functional_2009}
Kuhn, W.: A {Functional} {Ontology} of {Observation} and {Measurement}. In:
  Janowicz, K., Raubal, M., Levashkin, S. (eds.) {GeoSpatial} {Semantics}, pp.
  26--43. No. 5892 in Lecture {Notes} in {Computer} {Science}, Springer Berlin
  Heidelberg (2009)

\bibitem{prov}
Lebo, T., Sahoo, S., McGuinness, D., Belhajjame, K., Cheney, J., Corsar, D.,
  Garijo, D., Soiland-Reyes, S., Zednik, S., Zhao, J.: Prov-o: The prov
  ontology. W3C Recommendation, 30th April  (2013)

\bibitem{oboe}
Madin, J., Bowers, S., Schildhauer, M., Krivov, S., Pennington, D., Villa, F.:
  An ontology for describing and synthesizing ecological observation data.
  Ecological informatics  2(3),  279--296 (2007)

\bibitem{jefferson}
McGuinness, D.L., Pinheiro, P., Patton, E.W., Chastain, K.: In21b-3712 semantic
  escience for ecosystem understanding and monitoring: The jefferson project
  case study. In: Proceedings of AGU Fall Meeting 2014 (December 15-19 2014,
  Moscone Center, San Francisco, CA, US) (2014)

\bibitem{probst_ontological_2006}
Probst, F.: Ontological {Analysis} of {Observations} and {Measurements}. In:
  Raubal, M., Miller, H.J., Frank, A.U., Goodchild, M.F. (eds.) Geographic
  {Information} {Science}, pp. 304--320. No. 4197 in Lecture {Notes} in
  {Computer} {Science}, Springer Berlin Heidelberg (2006)

\bibitem{quine_stimulus_1995}
Quine, W.V.O.: From {Stimulus} to {Science}. Harvard University Press (1995)

\bibitem{santos_contextual_2015}
Santos, H., Pinheiro, P., McGuinness, D.L.: Contextual {Data} {Collection} for
  {Smart} {Cities}. In: Proceedings of the {Sixth} {Workshop} on {Semantics}
  for {Smarter} {Cities}. Bethlehem, PA, USA (2015)

\bibitem{stasch_stimulus-centric_2009}
Stasch, C., Janowicz, K., Bröring, A., Reis, I., Kuhn, W.: A
  {Stimulus}-{Centric} {Algebraic} {Approach} to {Sensors} and {Observations}.
  In: Trigoni, N., Markham, A., Nawaz, S. (eds.) {GeoSensor} {Networks}, pp.
  169--179. No. 5659 in Lecture {Notes} in {Computer} {Science}, Springer
  Berlin Heidelberg (2009)

\bibitem{stavropoulos_bonsai:_2012}
Stavropoulos, T.G., Vrakas, D., Vlachava, D., Bassiliades, N.: {BOnSAI}: {A}
  {Smart} {Building} {Ontology} for {Ambient} {Intelligence}. In: Proceedings
  of the 2Nd {International} {Conference} on {Web} {Intelligence}, {Mining} and
  {Semantics}. pp. 30:1--30:12. {WIMS} '12, ACM, New York, NY, USA (2012)

\bibitem{usbeck_combining_2014}
Usbeck, R.: Combining {Linked} {Data} and {Statistical} {Information}
  {Retrieval}. In: Presutti, V., d’Amato, C., Gandon, F., d’Aquin, M.,
  Staab, S., Tordai, A. (eds.) The {Semantic} {Web}: {Trends} and {Challenges},
  pp. 845--854. No. 8465 in Lecture {Notes} in {Computer} {Science}, Springer
  International Publishing (Jan 2014)

\end{thebibliography}

\end{document}